\documentclass{article}
\usepackage{ijcai24}

\usepackage{times}
\usepackage{soul}
\usepackage{url}
\usepackage[hidelinks]{hyperref}
\usepackage[utf8]{inputenc}
\usepackage[small]{caption}
\usepackage{graphicx}
\usepackage{amsmath}
\usepackage{amsthm}
\usepackage{booktabs}
\usepackage{algorithm}
\usepackage{algorithmic}
\usepackage[switch]{lineno}

\usepackage{graphicx}
\usepackage[utf8]{inputenc} 
\usepackage[T1]{fontenc}    
\usepackage{url}            
\usepackage{booktabs}       
\usepackage{amsfonts}       
\usepackage{nicefrac}       
\usepackage{microtype}      
\usepackage{epsfig}
\usepackage{float}
\usepackage{multicol}
\usepackage{multirow}
\usepackage{amssymb}
\usepackage{amsmath}
\usepackage{wrapfig}
\usepackage{xspace}
\usepackage{color}
\usepackage{colortbl}
\usepackage[dvipsnames, svgnames, x11names]{xcolor}
\usepackage[misc]{ifsym}
\usepackage{makecell}
\usepackage{soul}
\usepackage{bm}
\usepackage{wrapfig}

\definecolor{linkcolor}{RGB}{255,0,0}
\definecolor{urlcolor}{RGB}{255,105,180}
\definecolor{citecolor}{RGB}{66,168,235}

\usepackage{adjustbox}
\usepackage{pifont}
\usepackage{caption}
\usepackage{array}
\usepackage{enumitem}
\newcolumntype{C}[1]{>{\centering\arraybackslash}p{#1}} 

\def \pzo {\phantom{0}}

\definecolor{lightgray}{rgb}{0.8, 0.8, 0.8}
\definecolor{lgray}{rgb}{0.66, 0.66, 0.66}
\definecolor{whit_tab}{RGB}{255, 255, 255}
\definecolor{gray_tab}{RGB}{235, 235, 235}
\definecolor{oran_tab}{RGB}{254, 247, 241}
\definecolor{blue_tab}{RGB}{200, 227, 245}
\definecolor{lblu_tab}{RGB}{231, 239, 248}

\usepackage[capitalize]{cleveref}
\crefname{section}{Sec.}{Secs.}
\Crefname{section}{Section}{Sections}
\Crefname{table}{Table}{Tables}
\crefname{table}{Tab.}{Tabs.}

\newlength\savewidth

\renewcommand{\paragraph}[1]{\vspace{1.25mm}\noindent\textbf{#1}}

\newcommand{\ie}{i.e}
\newcommand{\eg}{e.g}

\def\onedot{.\xspace}
\def\eg{\emph{e.g}\onedot} 

\def\ie{\emph{i.e}\onedot}

\def\cf{\emph{c.f}\onedot}

\usepackage{tablefootnote}
\usepackage{diagbox}

\makeatother

\usepackage{adjustbox}

\title{GPT-4V-AD: Exploring Grounding Potential of \\VQA-oriented GPT-4V for Zero-shot Anomaly Detection}

\author{
Jiangning Zhang$^1$\footnotemark[1],
Haoyang He$^2$\footnotemark[1],
Xuhai Chen$^2$,
Zhucun Xue$^2$,
Yabiao Wang$^1$, 
Chengjie Wang$^1$ \\
Lei Xie$^2$\footnotemark[2], 
Yong Liu$^2$ \\
\affiliations
$^1$Youtu Lab, Tencent ~~ $^2$Zhejiang University
}

\begin{document}

\maketitle

\begin{abstract}
Large Multimodal Model (LMM) GPT-4V(ision) endows GPT-4 with visual grounding capabilities, making it possible to handle certain tasks through the Visual Question Answering (VQA) paradigm. This paper explores the potential of VQA-oriented GPT-4V in the recently popular visual Anomaly Detection (AD) and is the first to conduct qualitative and quantitative evaluations on the popular MVTec AD and VisA datasets. Considering that this task requires both image-/pixel-level evaluations, the proposed GPT-4V-AD framework contains three components: \textbf{\textit{1)}} Granular Region Division, \textbf{\textit{2)}} Prompt Designing, \textbf{\textit{3)}} Text2Segmentation for easy quantitative evaluation, and have made some different attempts for comparative analysis. The results show that GPT-4V can achieve certain results in the zero-shot AD task through a VQA paradigm, such as achieving image-level 77.1/88.0 and pixel-level 68.0/76.6 AU-ROCs on MVTec AD and VisA datasets, respectively. However, its performance still has a certain gap compared to the state-of-the-art zero-shot method, \eg, WinCLIP and CLIP-AD, and further researches are needed. This study provides a baseline reference for the research of VQA-oriented LMM in the zero-shot AD task, and we also post several possible future works. Code is available at \url{https://github.com/zhangzjn/GPT-4V-AD}.
\end{abstract}

\section{Introduction} \label{section:intro}
GPT-4V(ision)~\cite{openai_gpt4v_system_card} is a recent enhancement of GPT-4~\cite{openai_gpt4} released by OpenAI. It allows users to input additional images to extend the pure language model, implementing user interaction through a Visual Question Answering (VQA) manner. Recent works~\cite{yang2023dawn,yang2023set,wu2023early,shi2023exploring} have explored its potential in various settings, demonstrating its powerful generalization capabilities. On the other hand, due to the growing demand for industrial applications and the development of datasets~\cite{mvtec,visa,realiad}, Anomaly Detection (AD) is receiving increasing attention from researchers and practitioners~\cite{patchcore,rd,uniad,ocrgan,simplenet,memkd}. The recent zero-shot AD is first proposed in WinCLIP~\cite{winclip}, a setting dedicated to detecting image-level and pixel-level anomalies in a given image without any positive or negative samples. This setting addresses the pain point of difficulty in obtaining anomaly samples in industrial applications, thus having high practical value. Current approaches~\cite{winclip,aprilgan,ssa,anomalygpt,clipad} distinguishes anomalies based on a large language model, \eg, CLIP~\cite{clip}, which uses pre-trained vision language alignment to achieve anomaly detection. This framework often requires careful design. Unlike the aforementioned approach, this report explores a more general VQA paradigm for the zero-shot anomaly detection setting, hoping to bring new ideas to the solution of the zero-shot AD setting.

Specifically, this technical report explores the results of VQA-oriented LMM (using GPT-4V in this paper) on the zero-shot AD setting for the first time and proposes a general VQA-based AD framework, which includes \textbf{\textit{1)}} Granular Region Division (\cref{sec:method_image}), \textbf{\textit{2)}}Prompt Designing (\cref{sec:method_prompt}), and \textbf{\textit{3)}} Text2Segmentation (\cref{sec:method_t2s}). We conduct quantitative and qualitative experiments on the popular MVTec AD and VisA datasets (\cf, \cref{sec:mvtec}, \cref{sec:visa}, \cref{sec:qualitative_mvtec_visa}). The results show that GPT-4V has certain effects on the zero-shot AD setting, and even surpasses the zero-shot SoTA method in some metrics, such as achieving 88.0 AU-ROC on VisA, surpassing SoTA CLIP-AD by +6.8$\uparrow$. However, given that the anomaly detection setting requires pixel-level grounding capabilities, the current performance of GPT-4V still needs to be further improved. We hope this technical report can promote more zero-shot AD researches, especially the VQA-oriented paradigm.

\section{Related Work} \label{section:related}
\noindent\textbf{Visual Anomaly Detection.} Visual anomaly detection, classified by setting, encompasses full-shot unsupervised and zero-/few-shot approaches. 

Among these, full-shot unsupervised methods, the most widely recognized, can be further divided into three categories: Synthetic-based~\cite{draem,cutpaste,schluter2022natural,anomalydiffusion}, Feature Embedding-based~\cite{patchcore,padim,uninformedstudents,SPADE,differnet}, and Reconstruction-based~\cite{rd,omnial,invad,diad}. Synthetic-based methods enhance anomaly localization accuracy by artificially generating anomalies during training. Feature embedding techniques map normal samples into a compact feature space, where they are compared with test inputs in feature dimensions. Reconstruction-based methods learn the distribution of normal samples during training and reconstruct anomaly regions as normal in the testing phase, facilitating anomaly detection through comparison. All these methods necessitate extensive amounts of normal data for training and learning the typical sample distribution.

Recent advancements have enabled visual anomaly detection with minimal or zero sample data~\cite{graphcore,radford,saa}. PatchCore~\cite{patchcore}, for instance, constructs a subset of feature cores using a limited number of normal samples for comparison purposes. Notably, several recent studies~\cite{aprilgan,clipad,winclip,anomalyclip} on few-shot and zero-shot anomaly detection using the CLIP~\cite{clip} model have garnered significant attention, highlighting the potential of CLIP's visual-linguistic model. AnomalyGPT~\cite{anomalygpt}, leveraging a large-scale visual-linguistic model based on GPT, has demonstrated promising results in tasks with limited samples. On the other hand, GPT-4V~\cite{gpt4v} conducts simple text inference on zero- and single-sample scenarios using prompts, but lacks quantitative evaluation metrics.

\noindent\textbf{Large Vision-Language Models.} Recent advances have seen the proliferation of Large Visual-Language Models (LVLMs)~\cite{radford}, with notable examples such as BLIP-2~\cite{blip2}, which employs Q-Former for efficient multimodal alignment. InstructBLIP~\cite{instructblip} consists of a visual encoder, Q-Former, and a large language model (LLM). GPT-4~\cite{gpt4v}, the first model to integrate vision and language, showcases exceptional performance in both modalities, built on a Transformer architecture. MiniGPT-4~\cite{minigpt} is composed of pre-trained components like Vicuna~\cite{vicuna} from LLMs, ViT-G~\cite{vit} for vision, and Q-Former. LLaVA~\cite{llava} integrates LLaMA~\cite{llama} from pre-trained LLMs with the CLIP~\cite{clip} visual encoder, demonstrating strong capabilities in visual-language understanding and inference.

However, these LVLMs primarily generate language text outputs, necessitating additional modules for image localization tasks. This paper investigates the potential of GPT-4V in zero-shot visual anomaly detection and provides quantitative analysis, including anomaly localization metrics.
\section{Methodology} \label{sec:method}

\begin{figure*}[tp]
    \centering
    \includegraphics[width=1.0\linewidth]{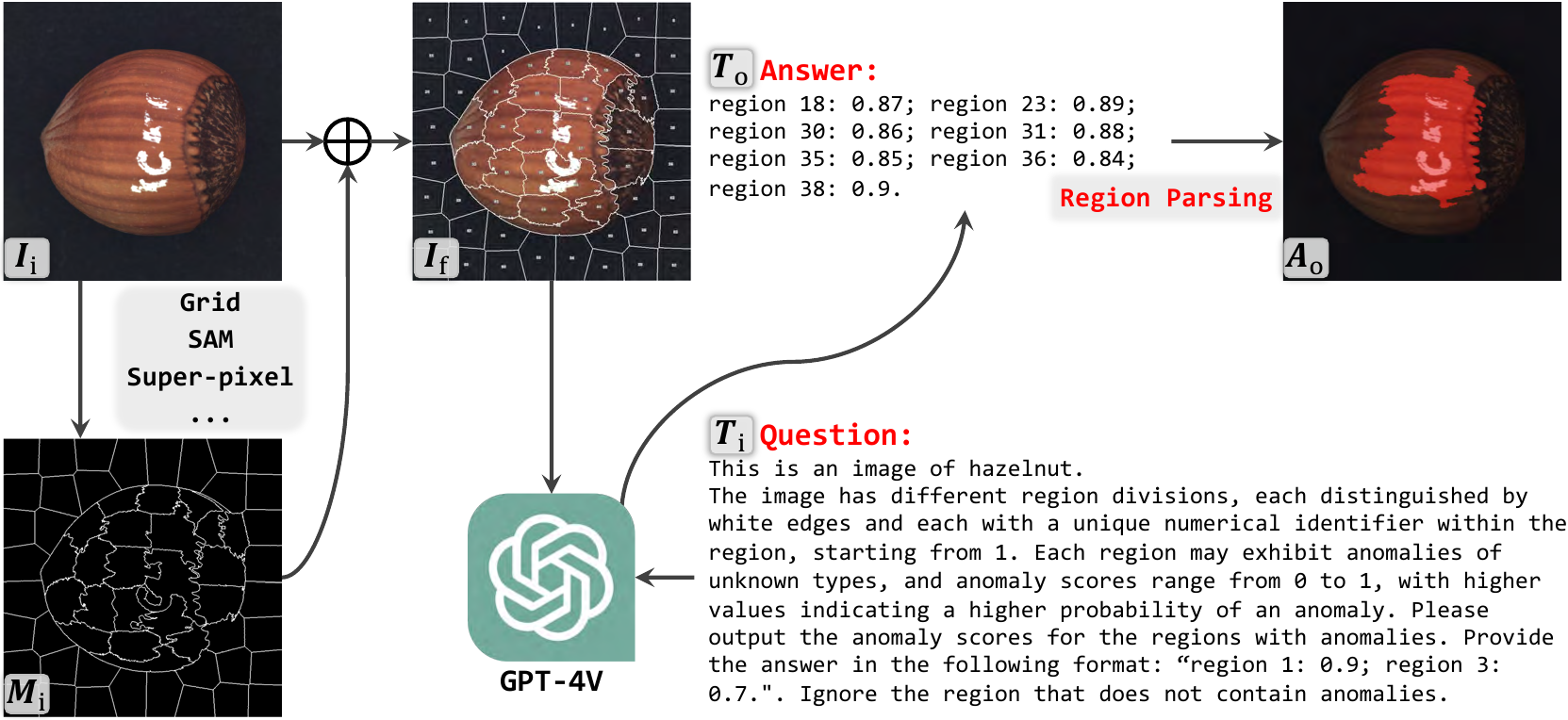}
    \caption{\textbf{Overview of the proposed GPT-4V-AD framework,} which consists of three procedures in tandem:  
    \textbf{\textit{1)}} \textbf{Granular Region Division} (\cref{sec:method_image}) preprocesses the input image $\bm{I}_{i}$, treating pixels that are similar at the structural or semantic level as a common region, resulting in $\bm{M}_{i}$. This is then combined with $\bm{I}_{i}$ through pixel-wise fusion to obtain the region-divided $\bm{I}_{f}$.
    \textbf{\textit{2)}} \textbf{Prompt Designing} (\cref{sec:method_prompt}) designs the suitable prompt $\bm{T}_{i}$ for the AD task in conjunction with $\bm{I}_{f}$, which is then input into GPT-4V to obtain a formatted output $\bm{T}_{o}$.
    \textbf{\textit{3)}} \textbf{Text2Segmentation} (\cref{sec:method_t2s}) combines the regions $\bm{M}_{i}$ to parse out pixel-level anomaly segmentation result $\bm{A}_{o}$.
    }
    \label{fig:method}
\end{figure*}

As a VQA model, LMM GPT-4V excels at comprehending input images at the semantic level but lacks pixel-level location perception. Anomaly detection tasks require pixel-level segmentation, so this section explores how to unleash the potential of GPT-4V's grounded vision-language capabilities in the AD task. Specifically, we propose a GPT-4V-AD framework that includes three components: Granular Region Division, Prompt Designing, and Text2Segmentation, as shown in \cref{fig:method}.

\begin{figure}[tp]
    \centering
    \includegraphics[width=1.0\linewidth]{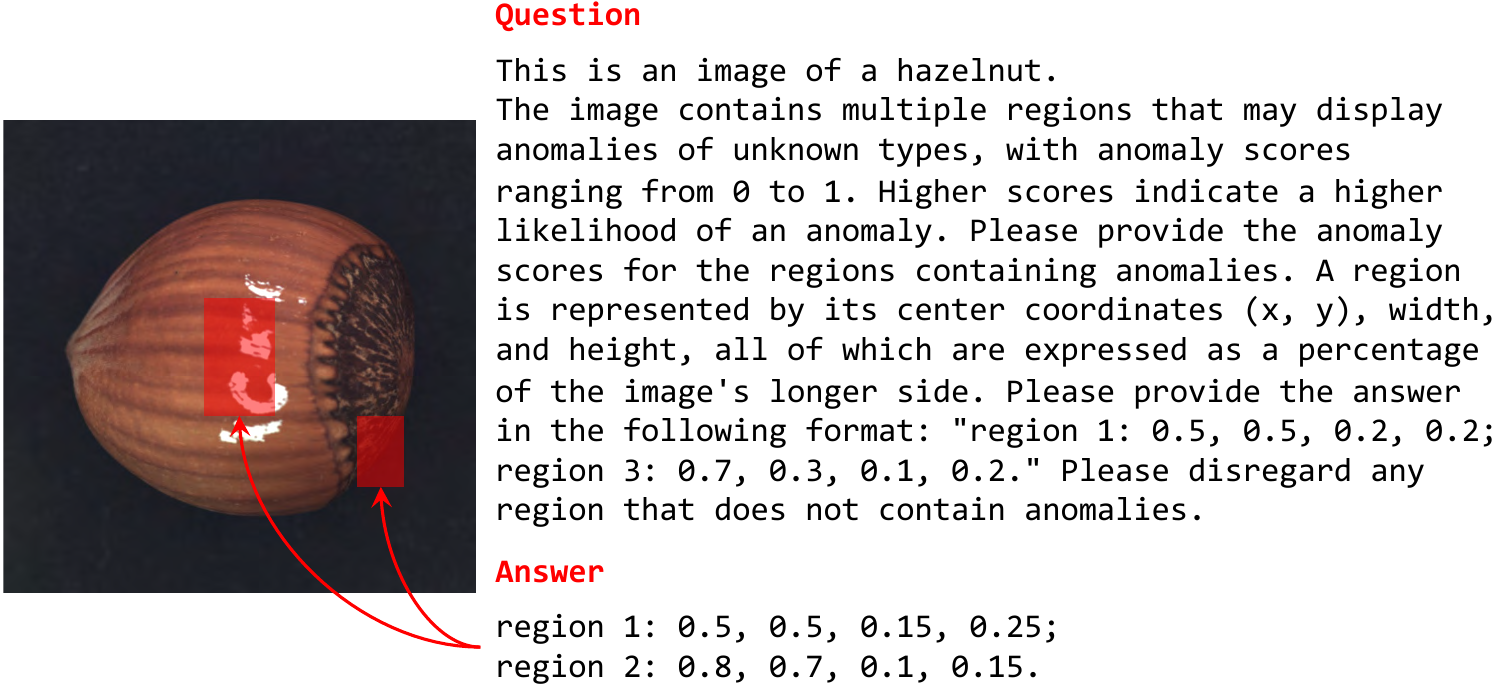}
    \caption{\textbf{A toy experiment with raw image as input and anomaly bounding box (expressed as percentage coordinates) as output for GPT-4V.} This manner leads to uncontrollable and imprecise outputs, and it is challenging to obtains pixel-level segmentation results.
    }
    \label{fig:toy}
\end{figure}

\begin{figure}[tp]
    \centering
    \includegraphics[width=1.0\linewidth]{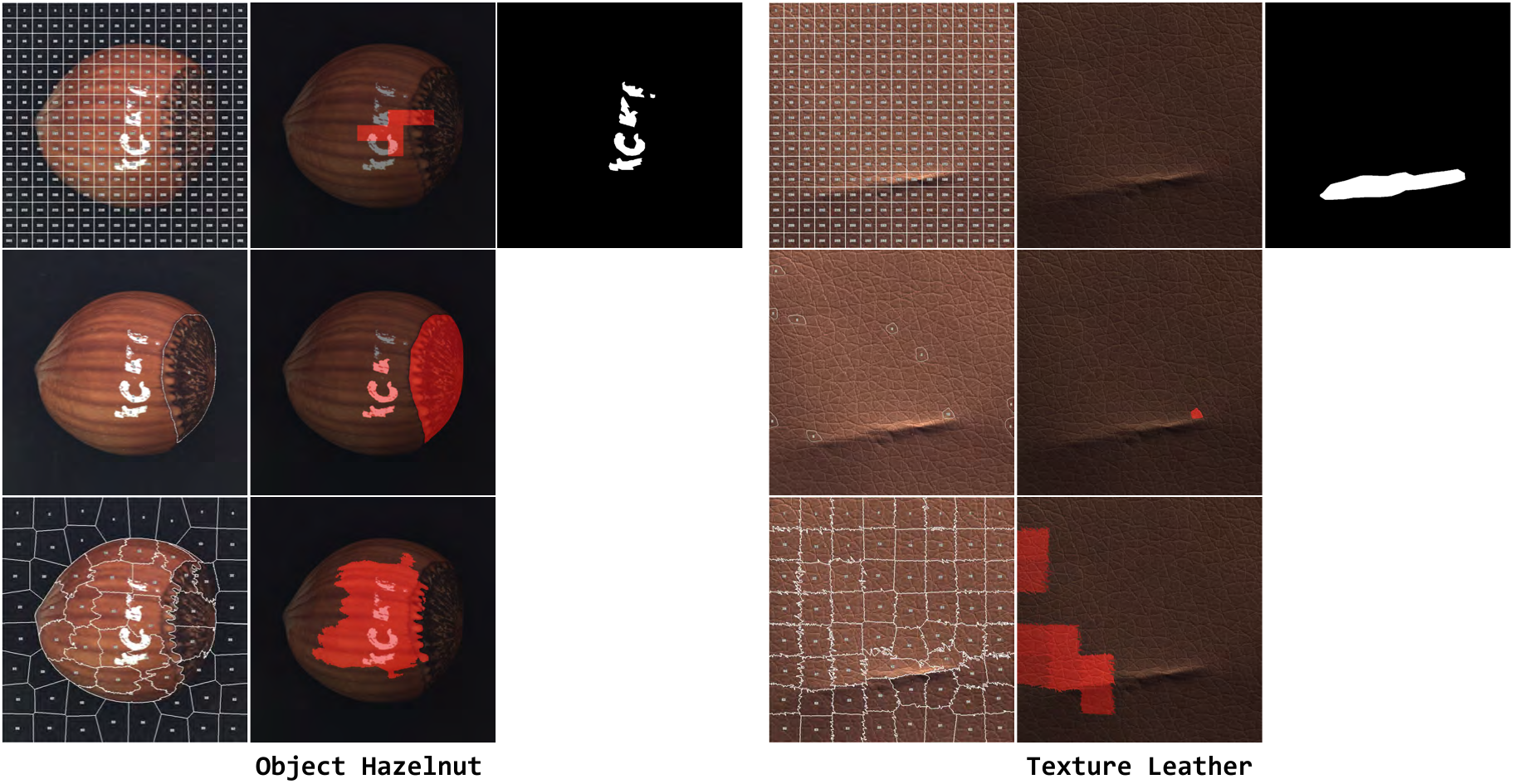}
    \caption{Ablation study on region division manners, \ie, naive gird, semantic SAM, and structural super-pixel.
    }
    \label{fig:region_division_comparison}
    \vspace{-1.5em}
\end{figure}

\subsection{Granular Region Division} \label{sec:method_image}
We first conduct a naive toy experiment. As shown in \cref{fig:toy}, when the original image and prompt are directly fed into GPT-4V, the model can generate judgments about defects, but cannot output accurate location results. We believe this is because GPT-4V can align text and object content at the semantic level, but is not adept at aligning text to pixel level. Therefore, \textit{we attempt to transform the AD problem based on VQA into a text and image region grounding problem}, which is more suitable for GPT-4V(ision). We assume that the anomalous regions have some form of connectivity under certain relationship constraints, such as semantics and local structure. The regions associated with the anomalous area should ideally closely adhere to the anomaly itself, ensuring that the segmentation results have a maximum upper limit.

\noindent\textbf{Region Generation.} \cref{fig:region_division_comparison} shows three region generation schemes we attempted: \textbf{\textit{1)}} Grid Sampling that uniformly samples different areas. \textbf{\textit{2)}} Semantic level region division generated by SAM~\cite{sam}. \textbf{\textit{3)}} Super-pixel manner that considers more similarity of layout structure. 
The qualitative results indicate that the grid manner does not pay enough attention to fine granularity, and the generated regions cannot cover the anomaly area well (\cf, first row of \cref{fig:region_division_comparison}); SAM may overkill by focusing on additional areas with obvious semantics (\cf, second row on the left), or discard areas with unclear semantics (\cf, second row on the right); Super-pixel can relatively stably generate divisions with a high overlap with anomalous ground truth. In addition, the three methods respectively achieve image-level 58.5/71.2/75.4 and 63.7/74.6/77.3 pixel-level AU-ROC results (limited by the number of GPT-4V accesses, experiments are only conducted on object hazelnut and texture leather). Results indicate that grid sampling, which has no structural prior, performs the worst, while the results of super-pixel are the best. Therefore, we recommend super-pixel as the default region generation module for the AD task.

\noindent\textbf{Region Labeling.} Similar to recent work~\cite{som}, the number is used as region labeling, but we outline each region in white without using a color filling. This is because anomaly detection is very sensitive to defect details, and this manner can minimize the impact of excessively small defect areas as illustrated in ~\cref{fig:method}.

\subsection{Prompt Designing} \label{sec:method_prompt}
Appropriate prompts are crucial for GPT-4V.~\cite{gpt4v} experimented with various prompt formats, evaluating both zero-shot and one-shot approaches for enhanced performance. For any test image of the popular AD datasets~\cite{mvtec,visa}, we can obtain the image category. Thus, we design a general prompt description for all categories and then inject the category information to it, \ie, $\bm{T}_{i}$ in~\cref{fig:method}. For example, '\textit{This is an image of hazelnut. The image has different region divisions, each distinguished by white edges and each with a unique numerical identifier within the region, starting from 1. Each region may exhibit anomalies of unknown types, and anomaly scores range from 0 to 1, with higher values indicating a higher probability of an anomaly. Please output the anomaly scores for the regions with anomalies. Provide the answer in the following format: “region 1: 0.9; region 3: 0.7.". Ignore the region that does not contain anomalies.}'

\subsection{Text2Segmentation} \label{sec:method_t2s}
Through structural output $\bm{T}_{o}$, we can easily obtain the final anomaly segmentation result $\bm{A}_{o}$ through regular matching combined with preprocessed regions $\bm{M}_{i}$. As illustrated in~\cref{fig:method}, the results corresponding to each region of $\bm{T}_{o}$ are populated into the corresponding areas of $\bm{M}_{i}$. Regions with higher anomaly scores are depicted in progressively deeper shades of red, while other areas remain black, indicating no detected anomalies. The resulting $\bm{A}_{o}$ constitutes the desired anomaly localization score map, which can be utilized for subsequent calculations and evaluations of quantitative metrics.

\section{Experiments} \label{sec:exp}

\subsection{Setup for Zero-shot AD} \label{sec:setup}
\noindent\textbf{Task Setting.} 
This investigation discusses the zero-shot AD task recently proposed in WinCLIP~\cite{winclip}, which aims to detect image-level and pixel-level anomalies without having seen samples of the anomalous categories. This setting is highly valuable for practical applications, as in many cases, anomalous samples are difficult to obtain or are extremely limited in quantity. Also, due to security reasons, the data may not be transferred externally. \textit{This paper explores the potential of GPT-4V, which is based on the VQA paradigm, in performing this task.}

\noindent\textbf{Dataset.} 
We evaluate GPT-4V(ision) with SoTAs on popular MVTec AD~\cite{mvtec} and VisA~\cite{visa} datasets for both anomaly classification and segmentation. In detail, MVTec AD contains 15 products in 2 types (\ie, object and texture) with 3,629 normal images for training and 467/1,258 normal/anomaly images for testing (5,354 images in total). VisA contains 12 objects in 3 types (\ie, single instance, multiple instance, and complex structure) with 8,659 normal images for training and 962/1,200 normal/anomaly images for testing (10,821 images in total). 

\begin{table}[tb]
    \caption{\textbf{Quantitative evaluation on MVTec AD dataset.} Top and middle parts shows .}
    The top and middle parts respectively show the single-category results of GPT-4V on texture and object. The bottom part shows the average results, as well as a comparison with recent SoTA zero-shot AD methods. The attempted VQA-oriented AD paradigm has achieved considerable results, but there is still a certain gap compared to the CLIP-based contrastive framework.
    \renewcommand{\arraystretch}{1.2}
    \setlength\tabcolsep{0.5pt}
    \resizebox{1.0\linewidth}{!}{
        \begin{tabular}{p{0.36cm}<{\centering} p{2.0cm}<{\centering} p{1.4cm}<{\centering} p{0.8cm}<{\centering} p{1.4cm}<{\centering} p{0.1cm}<{\centering} p{1.4cm}<{\centering} p{0.8cm}<{\centering} p{1.4cm}<{\centering} p{1.4cm}<{\centering}}
        \toprule[1.5pt]
        \multicolumn{2}{c}{\multirow{2}{*}{\makecell[c]{Category}}} & \multicolumn{3}{c}{Image-level} & & \multicolumn{4}{c}{Pixel-level} \\
        \cline{3-5} \cline{7-10}
        & & AU-ROC & AP & $F_1$-max & & AU-ROC & AP & $F_1$-max & AU-PRO\\
        \toprule[1.5pt]
        \multirow{5}{*}{\rotatebox{90}{\makecell[c]{Texture}}} & Carpet & 64.9 & 62.9 & 73.0 & & 69.9 & \pzo4.9 & 15.6 & 24.0 \\
        & Grid & 53.7 & 61.9 & 72.2 & & 60.7 & \pzo1.6 & \pzo5.0 & 28.6 \\
        & Leather & 69.3 & 62.8 & 70.3 & & 81.3 & \pzo5.5 & 12.8 & 56.8 \\
        & Tile & 94.1 & 93.8 & 88.8 & & 71.7 & 17.3 & 30.0 & 28.5 \\
        & Wood & 93.2 & 90.9 & 90.3 & & 67.8 & \pzo5.5 & 18.8 & 35.8 \\
        \hline
        \multirow{10}{*}{\rotatebox{90}{\makecell[c]{Object}}} & Bottle & 75.8 & 83.2 & 79.9 & & 56.2 & 11.3 & 20.6 & 24.4 \\
        & Cable & 77.9 & 77.7 & 71.7 & & 54.6 & \pzo4.6 & \pzo8.1 & \pzo8.9 \\
        & Capsule & 55.0 & 60.3 & 68.8 & & 63.5 & \pzo1.5 & \pzo2.6 & 38.9 \\
        & Hazelnut & 81.4 & 81.4 & 84.2 & & 73.3 & 10.8 & 24.4 & 50.2 \\
        & Metal Nut & 96.2 & 94.9 & 89.7 & & 52.6 & \pzo7.2 & 13.3 & 22.7 \\
        & Pill & 97.0 & 34.7 & 66.4 & & 83.5 & 12.9 & 29.9 & 30.9 \\
        & Screw & 99.0 & 38.0 & 66.1 & & 74.4 & \pzo1.5 & \pzo2.1 & 29.2 \\
        & Toothbrush & 75.2 & 75.5 & 67.4 & & 88.1 & \pzo2.5 & 13.1 & 60.7 \\
        & Transistor & 70.6 & 67.8 & 71.9 & & 56.5 & \pzo7.4 & 14.9 & 10.2 \\
        & Zipper & 53.4 & 62.0 & 66.4 & & 65.9 & \pzo1.9 & \pzo8.3 & 20.7 \\
        \hline
        \multirow{7}{*}{\rotatebox{90}{\makecell[c]{Zero-shot \\ SoTAs}}} & \multirow{2}{*}{\makecell[c]{Average \\(GPT-4V-AD)}} & \multirow{2}{*}{77.1} & \multirow{2}{*}{69.9} & \multirow{2}{*}{75.1} & & \multirow{2}{*}{68.0} & \multirow{2}{*}{\pzo6.4} & \multirow{2}{*}{14.6} & \multirow{2}{*}{31.4} \\
        & & & & & & & & & \\
        \cline{2-10}
        & WinCLIP & 91.8 & 96.5 & 92.9 & & 85.1 & - & 31.7 & 64.6 \\
        & SAA & 44.8 & 73.8 & 84.3 & & 67.7 & 15.2 & 23.8 & 31.9 \\
        & SAA+ & 63.1 & 81.4 & 87.0 & & 73.2 & 28.8 & 37.8 & 42.8 \\
        & CLIP-AD & 89.9 & 95.5 & 91.1 & & 88.7 & 28.5 & 35.3 & 89.9 \\
        & CLIP-AD+ & 90.8 & 95.4 & 91.4 & & 91.2 & 39.4 & 41.9 & 85.6 \\
        \bottomrule[1.5pt]
        \end{tabular}
    }
    \label{tab:mvtec}
\end{table}

\noindent\textbf{Metric.} 
Following prior works~\cite{winclip,aprilgan}, we use threshold-independent sorting metrics: \textit{\textbf{1)}} mean Area Under the Receiver Operating Curve (AU-ROC), \textit{\textbf{2)}} mean Average Precision~\cite{draem} (AP), and \textit{\textbf{3)}} mean $F_1$-score at optimal threshold~\cite{visa} ($F_1$-max) for both image-level and pixel-level evaluations. And \textit{\textbf{4)}} mean Area Under the Per-Region-Overlap~\cite{uninformedstudents} (AU-PRO) is also employed.

\noindent\textbf{Implementation Details.} 
The input image resolution is set to 768$\times$768 to maintain consistency with GPT-4V's input. In the region divisions, areas smaller than 600 or larger than 120K are filtered out. The edge of the region is outlined with a 1-pixel border, and the numerical labeling is placed within the mask and as centrally as possible within the entire region. For SAM~\cite{sam}, we use ViT-H~\cite{vit} as the region division backbone, and SLIC~\cite{slic} is chosen as the super-pixel approach with 60 segments and 20 compactness. 

\subsection{Quantitative Results on MVTec AD} \label{sec:mvtec}

We evaluate the zero-shot generalization ability of GPT-4V(ision) on the MVTec AD dataset~\cite{mvtec}. As shown in \cref{tab:mvtec}, the VQA-oriented framework also has category bias, \ie, it cannot maintain consistent performance across different categories, which is also reflected in CLIP-based contrastive zero-shot methods. The bottom of \cref{tab:mvtec} shows a comparison with the results of the recent zero-shot methods~\cite{winclip,ssa,clipad}. It can be seen that the VQA-oriented method has similar effects to the most recent SAA~\cite{ssa}, but there is still a certain gap compared to the SoTA results, which still needs further research.

\subsection{Quantitative Results on VisA} \label{sec:visa}

\begin{table}[tb]
    \caption{\textbf{Quantitative evaluation on VisA dataset.} .}
    The top three parts shows the results of GPT-4V on different categories. The bottom part shows the average results, as well as a comparison with recent SoTA zero-shot AD results. The attempted VQA-oriented AD paradigm has achieved highly competitive results on some metrics, but overall, there is still a certain gap compared to the CLIP-based contrastive framework.
    \renewcommand{\arraystretch}{1.2}
    \setlength\tabcolsep{0.5pt}
    \resizebox{1.0\linewidth}{!}{
        \begin{tabular}{p{0.36cm}<{\centering} p{2.3cm}<{\centering} p{1.4cm}<{\centering} p{0.8cm}<{\centering} p{1.4cm}<{\centering} p{0.1cm}<{\centering} p{1.4cm}<{\centering} p{0.8cm}<{\centering} p{1.4cm}<{\centering} p{1.4cm}<{\centering}}
        \toprule[1.5pt]
        \multicolumn{2}{c}{\multirow{2}{*}{\makecell[c]{Category}}} & \multicolumn{3}{c}{Image-level} & & \multicolumn{4}{c}{Pixel-level} \\
        \cline{3-5} \cline{7-10}
        & & AU-ROC & AP & $F_1$-max & & AU-ROC & AP & $F_1$-max & AU-PRO\\
        \toprule[1.5pt]
        \multirow{4}{*}{\rotatebox{90}{\makecell[c]{Complex \\Structure}}} & PCB1 & 100. & 37.7 & 65.8 & & 70.6 & \pzo1.3 & \pzo1.0 & \pzo8.0 \\
        & PCB2 & 100. & 37.0 & 65.8 & & 67.2 & \pzo1.5 & \pzo1.8 & \pzo6.2 \\
        & PCB3 & 92.7 & 85.8 & 91.3 & & 54.7 & \pzo0.6 & \pzo1.2 & \pzo0.3 \\
        & PCB4 & 73.3 & 71.9 & 77.6 & & 96.7 & 10.5 & 12.3 & 78.3 \\
        \hline
        \multirow{4}{*}{\rotatebox{90}{\makecell[c]{Multiple \\Instances}}} & Macaroni1 & 95.9 & 94.3 & 89.8 & & 98.7 & \pzo1.8 & \pzo2.8 & 34.2 \\
        & Macaroni2 & 69.2 & 44.4 & 66.3 & & 76.6 & \pzo1.0 & \pzo0.6 & 41.3 \\
        & Capsules & 87.2 & 38.2 & 67.2 & & 90.5 & \pzo0.9 & \pzo1.8 & 36.8 \\
        & Candle & 99.2 & 100 & 100 & & 54.8 & \pzo1.1 & \pzo0.7 & 30.8 \\
        \hline
        \multirow{4}{*}{\rotatebox{90}{\makecell[c]{Single \\Instance}}} & Cashew & 54.0 & 56.0 & 67.2 & & 57.0 & \pzo4.3 & 12.8 & \pzo6.3 \\
        & Chewing Gum & 85.4 & 85.1 & 76.2 & & 74.8 & 12.1 & 31.1 & 11.9 \\
        & Fryum & 99.2 & 99.5 & 99.9 & & 81.4 & 15.9 & 23.2 & 28.7 \\
        & Pipe Fryum & 99.5 & 99.1 & 100. & & 96.0 & \pzo0.1 & \pzo0.9 & 71 \\
        \hline
        \multirow{7}{*}{\rotatebox{90}{\makecell[c]{Zero-shot \\ SoTAs}}} & \multirow{2}{*}{\makecell[c]{Average \\(GPT-4V-AD)}} & \multirow{2}{*}{88.0} & \multirow{2}{*}{70.7} & \multirow{2}{*}{80.6} & & \multirow{2}{*}{76.6} & \multirow{2}{*}{\pzo4.3} & \multirow{2}{*}{\pzo7.5} & \multirow{2}{*}{29.5} \\
        & & & & & & & & & \\
        \cline{2-10}
        & WinCLIP & 78.1 & 81.2 & 79.0 & & 79.6 & - & 14.8 & 56.8 \\
        & SAA & 48.5 & 60.3 & 73.1 & & 83.7 & \pzo5.5 & 12.8 & 41.9 \\
        & SAA+ & 71.1 & 77.3 & 76.2 & & 74.0 & 22.4 & 27.1 & 36.8 \\
        & CLIP-AD & 81.2 & 83.7 & 80.0 & & 84.1 & \pzo9.6 & 16.0 & 63.4 \\
        & CLIP-AD+ & 81.1 & 85.1 & 80.9 & & 94.8 & 20.3 & 26.5 & 85.3 \\
        \bottomrule[1.5pt]
        \end{tabular}
    }
    \label{tab:visa}
\end{table}

We further evaluate the performance of GPT-4V(ision) on the popular VisA dataset~\cite{visa}, which contains more small defects and is more challenging. Surprisingly, unlike the MVTec AD results, the proposed VQA paradigm performs significantly better on this dataset and even surpasses the SoTA method on some metrics, such as achieving an image-level AU-ROC of 88.0 that surpasses SoTA CLIP-AD by +6.8$\uparrow$.

\subsection{Qualitative results} \label{sec:qualitative_mvtec_visa}

\begin{figure*}[tp]
    \centering
    \includegraphics[width=1.0\linewidth]{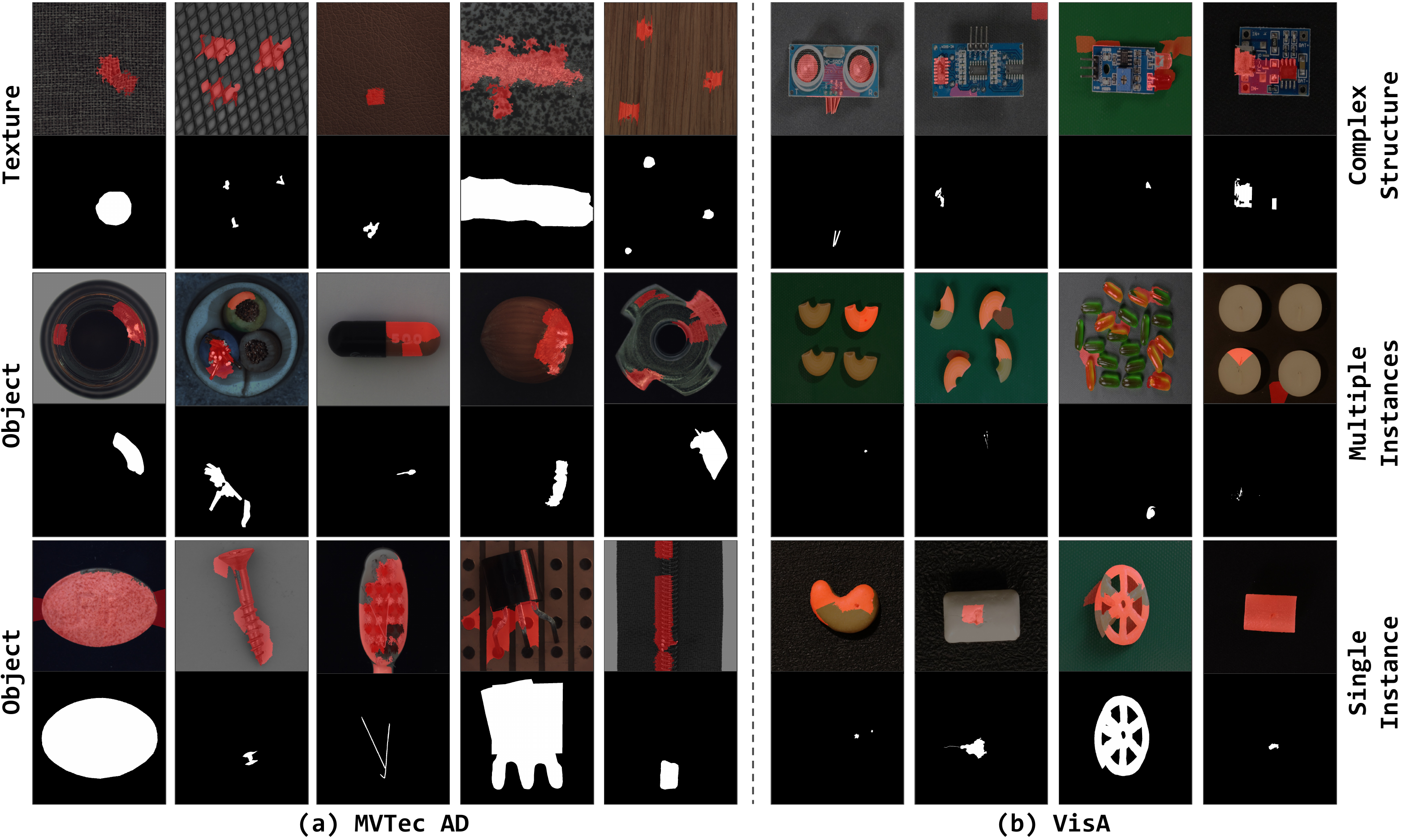}
    \caption{Non-cherry-picked qualitative results for each category on the MVTec AD (left) and VisA (right) datasets. 
    }
    \label{fig:vis}
\end{figure*}

\begin{figure*}[tp]
    \centering
    \includegraphics[width=1\linewidth]{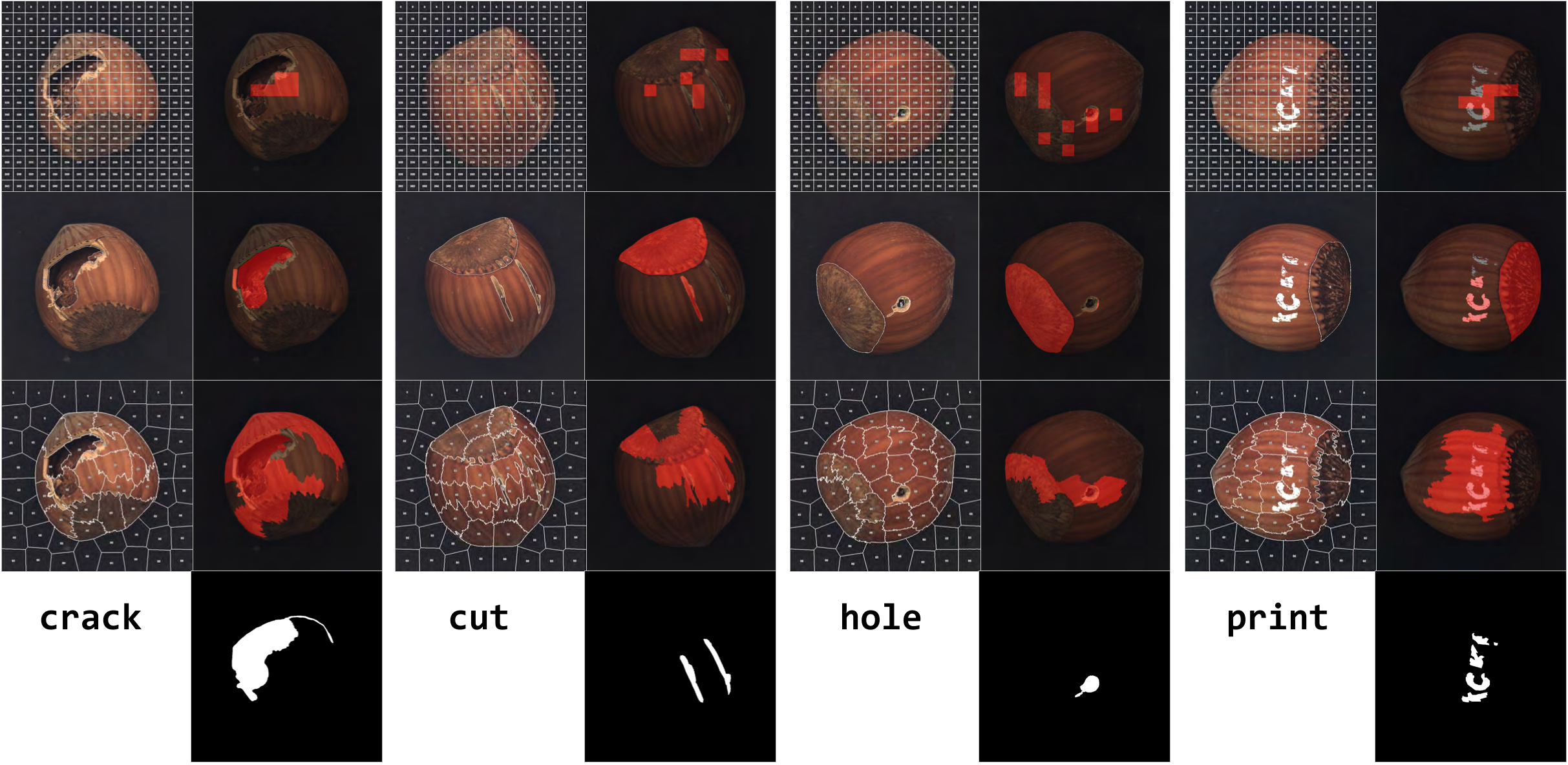}
    \caption{Qualitative result comparison for different defect categories in the \textbf{object hazelnut} on MVTec AD dataset.
    }
    \label{fig:vis_hazelnut}
\end{figure*}

\begin{figure*}[tp]
    \centering
    \includegraphics[width=1.0\linewidth]{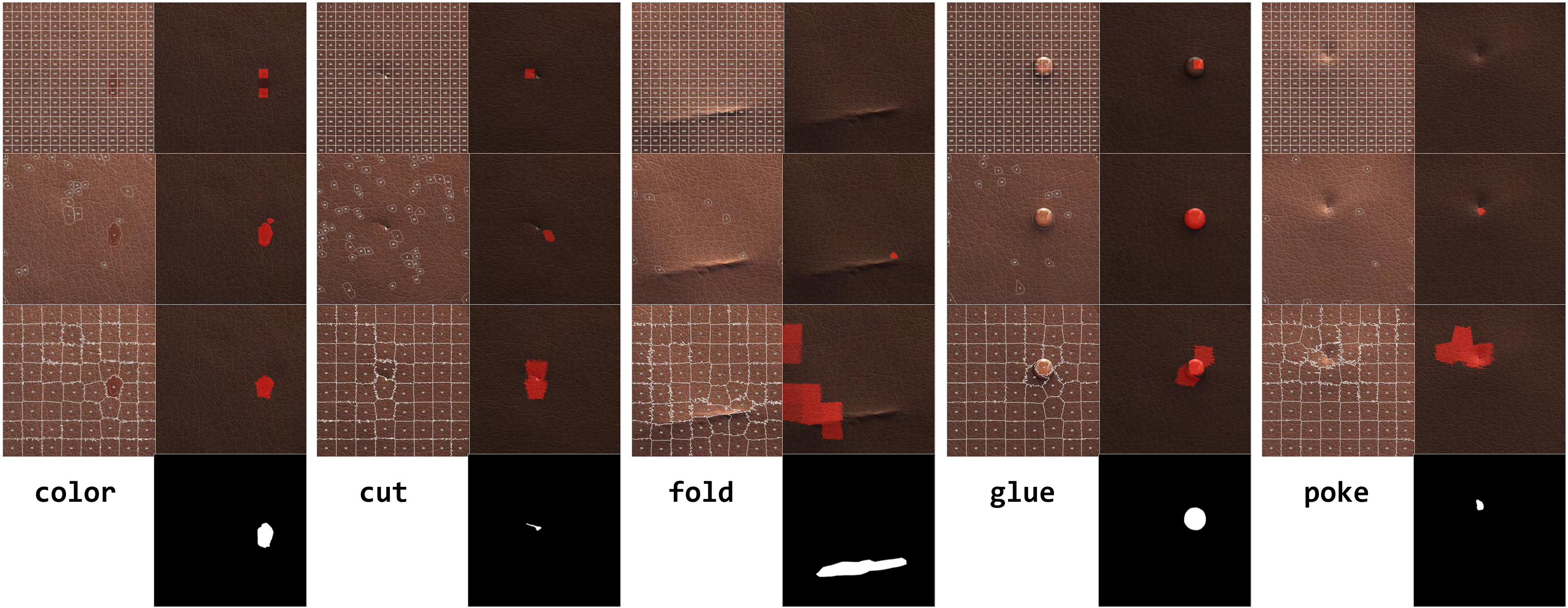}
    \caption{Qualitative result comparison for different defect categories in the \textbf{texture leather} on MVTec AD dataset.
    }
    \label{fig:vis_leather}
\end{figure*}

\cref{fig:vis} shows the qualitative segmentation results. Even based on the intuitive VQA approach, general-purpose GPT-4V can still obtain segmentation results closer to the ground truth of the anomaly region, demonstrating its powerful visual grounding ability.

\subsection{Analyses and Visualization} \label{sec:analyses_vis}



\noindent \textbf{1) Ablation study on region division manner.} 
This paper adopts three Granular Region Division methods, \ie, grid, SAM~\cite{sam}, and super-pixel~\cite{slic}. \cref{fig:vis_hazelnut} and \cref{fig:vis_leather} respectively show the qualitative results of the three methods on object hazelnut and texture leather. SAM tends to locate more semantically biased areas, ignoring some weak semantic areas with unclear edges, which has a good effect on defects with obvious semantic classification, \eg, crack in the hazelnut and glue in the leather. However, it may misjudge areas that are too semantically obvious (\cf, bottom of the hazelnut), and miss minor defects (\cf, fold defect in leather). In contrast, super-pixel can pay attention to every area in the image and provide better division results for partial structures than the grid manner.

\begin{figure}[tp]
    \centering
    \includegraphics[width=1.0\linewidth]{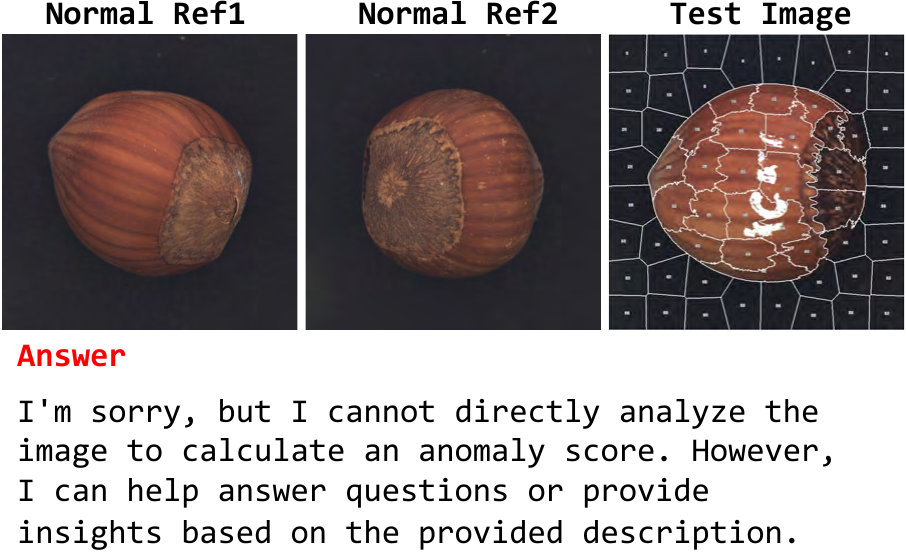}
    \caption{Analysis of results with extra normal reference images, but GPT-4V often fails to respond as expected.
    }
    \label{fig:ref_inputs}
\end{figure}

\noindent \textbf{2) Results with Extra Reference Images.} 
The few-shot setting introduces additional images into the model to improve its performance. Therefore, we further input two extra normal images as reference images into the model. The experimental results in \cref{fig:ref_inputs} show that the current version of GPT-4V cannot effectively utilize the additional reference images in the anomaly detection task, but often gets disturbed and cannot output results normally.

\begin{figure}[tp]
    \centering
    \includegraphics[width=1.0\linewidth]{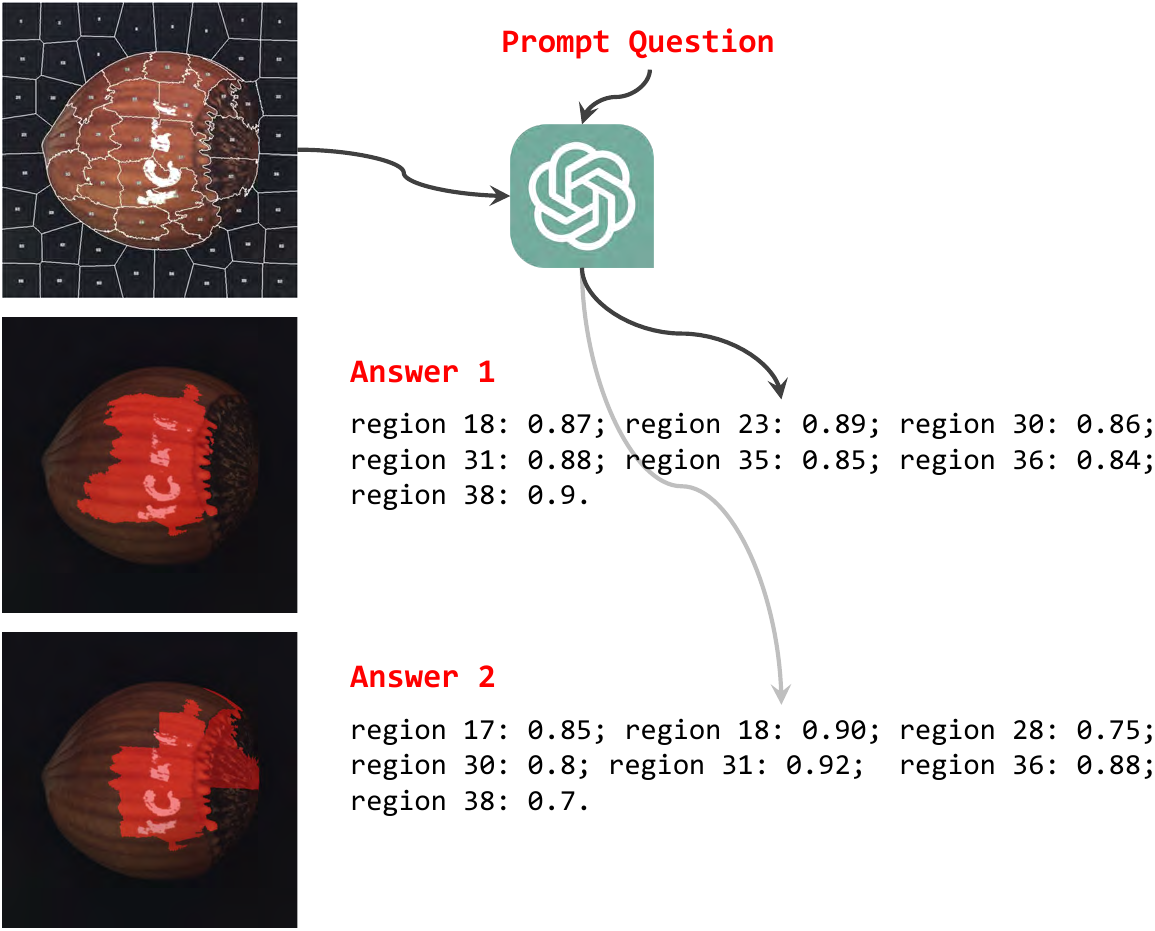}
    \caption{Stability analysis of VQA-oriented GPT-4V's output with multiple inputs, and there is a slight difference in the results as each time the same prompt question.
    }
    \label{fig:multiple_output}
\end{figure}

\noindent \textbf{3) Repeatability Analysis.} 
We conduct repeated experiments on the consistency of GPT-4V's results. As shown in \cref{fig:multiple_output}, when we use the same image and prompt inputs, the output anomaly regions will have slight differences, including both region number and confidence score.

\section{Conclusion and Future Works} \label{section:con}
This paper explores the potential of VQA-oriented GPT-4V(ision) in the zero-shot AD task and proposes adapted image and prompt processing methods for quantitative and qualitative evaluation. The results indicate that it has a certain effectiveness on popular AD datasets. Nevertheless, there is still room for further improvement in AD tasks. Moreover, although GPT-4V has achieved epoch-making improvements in human-machine interaction, how to better apply this capability to pixel-level fine-grained tasks remains to be further studied, and the issue of high inference cost needs to be addressed.

\noindent\textbf{Limitations and Future Works.}
We have summarized some of the current challenges and future works: 

\noindent\textbf{\textit{1)}} Due to the limitation on the number of accesses, more suitable image preprocessing methods and more complex prompt designs can be attempted to fully evaluate this task. 

\noindent\textbf{\textit{2)}} AD datasets are generally collected from specific scenarios, such as industry, and GPT-4V may have less data from these scenarios in its training set, which could lead to poor generalization performance. Therefore, specific fine-tuning for AD tasks can be studied. 

\noindent\textbf{\textit{3)}} This paper only explores the experimental results of VQA-based pure LMM (Large Multi-modal Models), and combining it with current zero-shot AD methods may further improve the model's performance.

\noindent\textbf{\textit{4)}} Prior learning of few-shot normal/anomalous samples should help GPT-4V better understand defects and grounding, which researchers can further explore. 

\noindent\textbf{\textit{5)}} Using GPT-4V for semi-automatic data annotation can reduce the cost of manual annotation. 

\noindent\textbf{\textit{6)}} The labeling scheme chosen may impact overly small defects, leading to overkill or missed detection. It would be worthwhile to explore more reasonable alternative solutions.

\noindent\textbf{\textit{7)}} The uniqueness of the output from the current paradigm is relatively poor, and there may even be significant gaps among different tests. This paper only provides a preliminary report, and further experiments are warranted for more stable model testing in the future.

\bibliographystyle{named}
\bibliography{ijcai24}

\end{document}